\documentclass[conference]{IEEEtran}
\IEEEoverridecommandlockouts
\usepackage[draft]{minted}
\usepackage{listings}
\usepackage{minted}
\usepackage{underscore}
\usepackage{amsmath,amssymb,amsfonts}
\usepackage{algorithmic}
\usepackage{graphicx}
\usepackage{textcomp}
\usepackage{comment}
\usepackage{xcolor}
\usepackage{multirow}
\usepackage[utf8]{inputenc}
\def\BibTeX{{\rm B\kern-.05em{\sc i\kern-.025em b}\kern-.08em
    T\kern-.1667em\lower.7ex\hbox{E}\kern-.125emX}}
\begin{document}
\title{Data Collection and Utilization Framework for Edge AI Applications}
\pagenumbering{arabic}

\author{\IEEEauthorblockN{Hergys Rexha, Sébastien Lafond}
\IEEEauthorblockA{\textit{Åbo Akademi University} \\
hergys.rexha@abo.fi, sebastien.lafond@abo.fi}

}
\IEEEaftertitletext{\vspace{-1\baselineskip}}
\maketitle  
\thispagestyle{plain}
\pagestyle{plain}

\begin{abstract}

 As data being produced by IoT applications continues to explode, there is a growing need to bring computing power closer to the source of the data to meet the response-time, power dissipation and cost goals of performance-critical applications in various domains like Industrial Internet of Things (IIoT), Automated Driving, Medical Imaging or Surveillance among others. This paper proposes a data collection and utilization framework that allows runtime platform and application data to be sent to an edge and cloud system via data collection agents running close to the platform. Agents are connected to a cloud system able to train AI models to improve overall energy efficiency of an AI application executed on a edge platform. In the implementation part we show the benefits of FPGA-based platform for the task of object detection. Furthermore we show that it is feasible to collect relevant data from an FPGA platform, transmit the data to a cloud system for processing and receiving feedback actions to execute an edge AI application energy efficiently. As future work we foresee the possibility to train, deploy and continuously improve a base model able to efficiently adapt the execution of edge applications.
\end{abstract}

\pagenumbering{arabic}

\vspace{-5pt}

\section{Introduction}
Edge computing is a fast-growing technology trend, which involves pushing compute capabilities to the edge. Edge computing can be described as a distributed computing paradigm that brings computation and data storage closer to the location needed to improve response times, save bandwidth, and improve security.

Edge systems are the deterministic embedded communication and real-time control engines that reside at the edge of the network and closest to the physical world of factories and other industrial environments, e.g., motion controllers, protection relays, programmable logic controllers, and similar systems. Clock frequencies in gigahertz, larger memory sizes, higher numbers of input/output ports, and the latest encryption engines might seem to offer solutions for future requirements. However, when dealing with the timescale of industrial equipment, which has critical subsystems that operate on a scale of hundreds of microseconds (or less) and need to operate in factories and remote locations for decades, relying solely on a cutting-edge multicore embedded processor is risky. %
A much higher degree of freedom in scaling is desperately needed, at for example Industrial edge system, due to the timescales involved. 
Also there is a need for a more consistent approach that allows determinism, latency, and performance to be easily managed.
At the heart of the current industrial revolution is the roll-out of machine learning (ML) algorithms, specifically deep neural networks (DNNs). They achieve impressive results in computer vision and speech recognition, and are increasingly being adopted for other tasks. DNNs are first trained on a labeled dataset, and afterwards can be used for inference on previously unseen data as part of an application. 
The large compute and storage requirements associated with DNN deployment necessitate acceleration. Furthermore, different constraints might be imposed on accuracy, cost, power, model size, throughput, and latency depending on the use case. Real-time and safety-critical applications such as augmented reality, drone control, and autonomous driving are not suitable for offloading to the cloud due to low-latency requirements and data transmission overhead. In cloud-computing and ML-as-a-Service contexts, data centers face ever-increasing throughput requirements to process astronomical scales of data~\cite{alibaba}, bringing additional challenges in energy efficiency to minimize operating expenses. While cloud service latency is less critical compared to embedded scenarios, it still translates directly into customer experience for interactive applications.
Traditionally, machine learning research was focused on improving the accuracy of the models without particular regard to the cost of inference. This is evident in the older networks like AlexNet and VGG, which are now considered large and with many parameters~\cite{Eyeriss}.
However, as machine learning and DNNs move into practical applications, compute and memory requirements become a major concern.

\begin{figure}[t]
    \centering
    \includegraphics[width=\columnwidth]{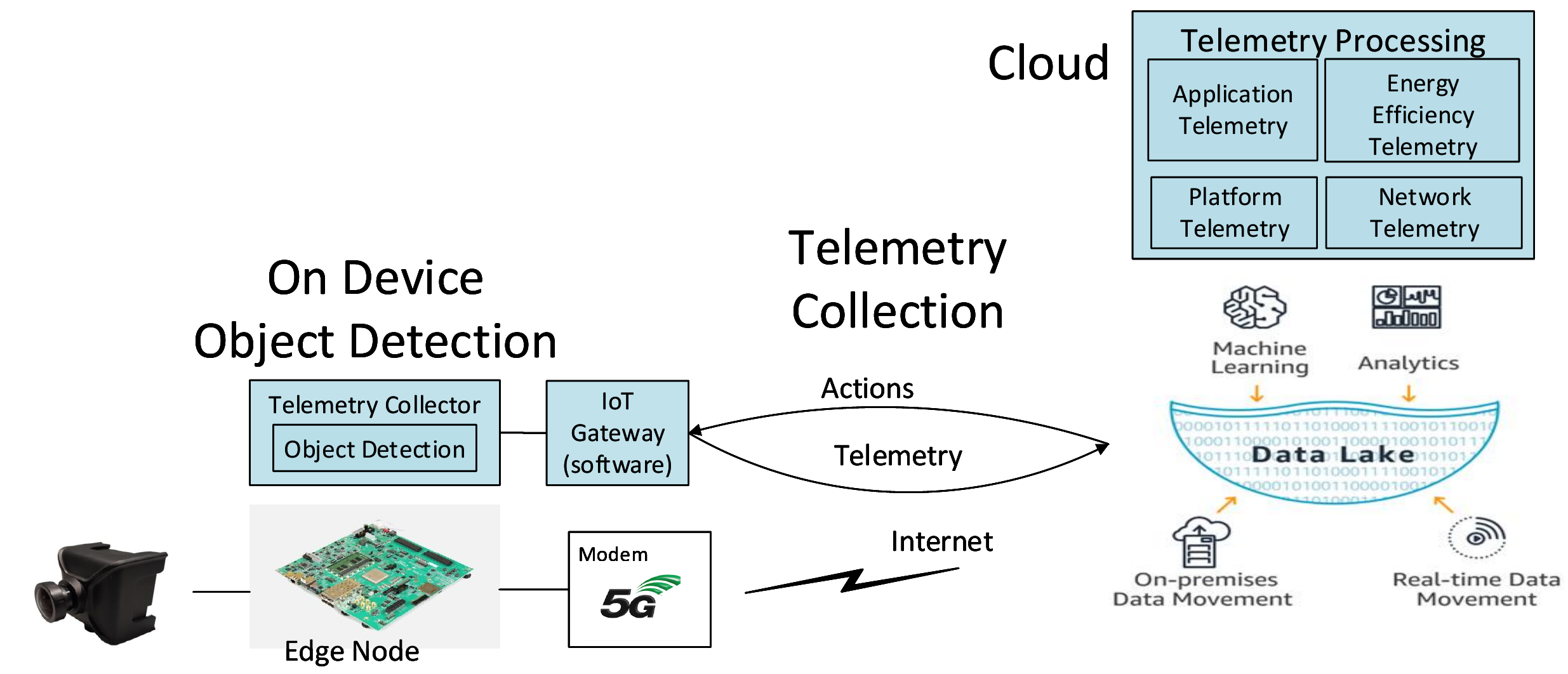}
    \caption{Basic schematic of the telemetry framework}
    \label{fig:telemetry-framework}
    \vspace{-6mm}
\end{figure}

\section{Use case and Architecture Overview}
The assumed scenario for this work is the following: an industrial system (it could be for example a patrolling robot or a manufacturing conveyor) is streaming live a video over a 5G network and requests as a service detected objects from the video stream. The object detection service is executed from the Multi-access Edge Computing (MEC) of the used 5G base station.

The main assumptions of the work are the following:
(a) an FPGA platform can provide lower latency than  more traditionally used GPU platforms for this type of application. This is due to the datapath architecture of the FPGA and DPU, which does not require to first “flood” a large number of Streaming Multiprocessors (SM) as in a GPU;
(b) taking advantage of the DPUs, we can reach higher throughputs in terms of number of processed frames per second;
(c) the FPGA platform will provide a better energy efficiency solution compared to CPU and GPU based solutions.

The main goals of the work are the following:
\begin{itemize}
     \item Propose and implement a telemetry collection framework that complements the use case scenario described above.
     \item Evaluate achievable latency, throughput and energy efficiency of common edge platform alternatives for the selected use case and proposed architecture.
     \item Show the benefits off customizing computations at the edge with the intelligence from the cloud side created with the collected telemetry data.

\end{itemize}

\subsection{Telemetry Framework}
There is a need for big data analytics and machine-learning based AI technologies for the operational automation of factories and other industrial environments. These use cases deploy edge systems for real-time control of the operations. The collection of large amounts of data is required from different system components like applications, edge platform and network. The single-sourced and static data acquisition cannot meet this data requirements. It is therefore desirable to have a framework that integrates multiple telemetry approaches from different components. The telemetry framework brings a solution to this problem.
The main focus of this work is to provide an end-to-end description and evaluation of the  proposed telemetry architecture which is described in Figure~\ref{fig:telemetry-framework}. The framework can be divided in two parts: the edge part which is described on the left of the Figure~\ref{fig:telemetry-framework}, and the cloud part which is on the right side.  At the edge side of the schematic we have a highly heterogeneous platform which is equipped, either with GPU or with re-configurable hardware (FPGA). The platform is hosting an intelligent application which uses a convolutional neural network (CNN) for performing real-time video inference, and an agent which is collecting several metrics from the application, platform, and network called the telemetry agent. Metrics of various components of the platform are collected and formatted as a JSON object and sent to the other part of the framework. On the cloud part the data is analyzed and actions are taken as a feedback controlling the behaviour of the intelligence performed on the edge side.    
A more detailed description of the telemetry framework is available from~\cite{rexha2021data}. 

\section{Edge platform technologies}
To support our assumptions on the achievable latency and performance of FPGA platforms for CNN-based edge applications, we evaluate two different platforms for the role of the edge node: a Nvidia Jetson AGX Xavier (as a representative GPU-based platform) and a Xilinx ZCU102 (as a representative FPGA-based platform). The Xavier is an embedded GPU platform which promise to offer high compute density and good energy efficiency for AI related applications. The Xavier is equipped with 512 CUDA cores with Volta architecture GPU running at 1.37GHz and a 16GB LPDDR4X @ 2133MHz memory with a bandwidth of 137 GB/s, and a flash storage eMMC 32GB. 
The Xilinx Zynq UltraScale+ MPSoC ZCU102 board has a 16nm XCZU9EG
FPGA, an on-board 4GB 64bit DDR4 RAM with a peak bandwidth of 136Gb/s.

We aim at testing the  AI inference capabilities, and the power dissipation of the two platforms  while running neural network algorithms. The experiments are conducted using Yolov3 and SSDResnet50Fpn algorithms for object detection which perform inference on a  420p video file. In Table \ref{table1} we report the measurements done for both platforms for metrics such as end-to-end delay (EE latency) to process a single frame and number of frames per second processed for a single dissipated watt (FPS/Watt) while running two popular object detection algorithms such as Yolo and SSD. The neural network is fed with the same video file and the power is measured on the entire platform. FPGA architecture is able to achieve good latency in time-sensitive jobs due to the circuit-level customizations on its massively parallel computing units. 
\begin{table}[h!]
\begin{center}
\begin{tabular}{ |c|c|c|c| } 
 \hline
 Platform & Algorithm & EE latency (ms) & FPS/Watt \\ 
 \hline
 \multirow{2}{4em}{Xavier AGX} & Yolov3 & 120 & 0,3 \\\cline{2-4}
 
 & SSD\_Resnet50\_fpn & 250  & 0,17 \\ 
 \hline
 \multirow{2}{4em}{ZCU102} &  Yolov3 & 29,4 & 1,48 \\ \cline{2-4}
 & SSD\_Resnet50\_fpn & 200  & 0,37 \\ 
 \hline
 \end{tabular}
\end{center}
\caption{Inference characteristics of the considered edge platforms}
\label{table1}
\vspace{-5mm}
\end{table}
From the results shown on the table there is a clear advantage of the FPGAs platforms versus GPUs to be used especially in streaming applications, this is noticeable in terms of latency and energy-efficiency. The SSD\_Resnet\_50\_FPN is a heavier model compared to Yolo, requiring 178.4 Gops compared to 65.63 Gops of the other side. Based on this evaluation, the ZCU102 FPGA-based board will be used a the edge device in the rest of the paper.

\section{Experimentation Methodology}

As described in section II the telemetry framework consist mainly in two parts: the Edge and Cloud side.
\subsection{Edge Side}
 On the edge side we are executing a CNN-based video inference application which is quantized and pruned for running on a FPGA device. We also collect the parameters which will make up our telemetry data through an agent that is running on the device. The agent collects telemetry data from three categories as described below:  
\subsubsection{Application Telemetry}
Latency of the application, FPS.
\subsubsection{Model Efficiency Telemetry}
Computational Unit utilization, Memory Throughput, CPU utilization, Memory utilization, AI model efficiency. 

\subsubsection{Energy Efficiency Telemetry}
Dissipated power, Temperature of the module, FPS/Watt.
\begin{comment}

\begin{itemize}
    \item Run-time power dissipation of the platform in Watt:
    \begin{itemize}
        \item Processing System Power dissipation. This includes Low Power Domain (LPD), Full Power Domain (FPD), and DDR Mem. Controller.
        \item Programmable Logic Power dissipation. This includes the Internal Power, and the Block RAM (BRAM) power.
        \item I/O power dissipation.
        \end{itemize}
    \item Temperature of the Programmable Logic in $^\circ$C
    \item Temperature of the Processing System in $^\circ$C
    \item Application performance-per-watt (FPS/Watt)
    \[\var{FPS/Watt} = \frac{\var{model_fps}}{\var{platform_power}} \]
\end{itemize}
 where \var{model_fps} represents the performance of the application in frame per second and \var{platform_power} represent the total power dissipation of the FPGA platform.
\end{comment}

\subsubsection{Communication Network Telemetry} 
From the network side the agent collects parameters of the 4G modem used in the experimental framework and these telemetry is sent transparently to the cloud. The parameters collected include: RSSI, RSRQ, RSRP, Temperature, DL/UP.
\begin{comment}

\begin{itemize}
    \item Received Signal Strength Indicator (RSSI). Received signal strength indicator (RSSI) measured in dBm. Values closer to 0 indicate a better signal strength.
    \item Reference Signal Received Quality (RSRQ). Quality considering also RSSI and the number of used Resource Blocks (N) RSRQ = (N * RSRP) / RSSI measured over the same bandwidth.
    \item Reference Signal Received Power (RSRP). It is the power of the LTE Reference Signals spread over the full bandwidth and narrowband.
    \item Signal to Interference plus Noise Ratio (SINR). Is a quantity used to give theoretical upper bounds on channel capacity (or the rate of information transfer) in wireless communication.
    \item Temperature
    \item Cell\_ID
    \item Download Rate (Mbps)
    \item Upload Rate (Mbps)
   
\end{itemize}
\end{comment}

 \begin{figure}[t]
\centering
\includegraphics[width=\columnwidth]{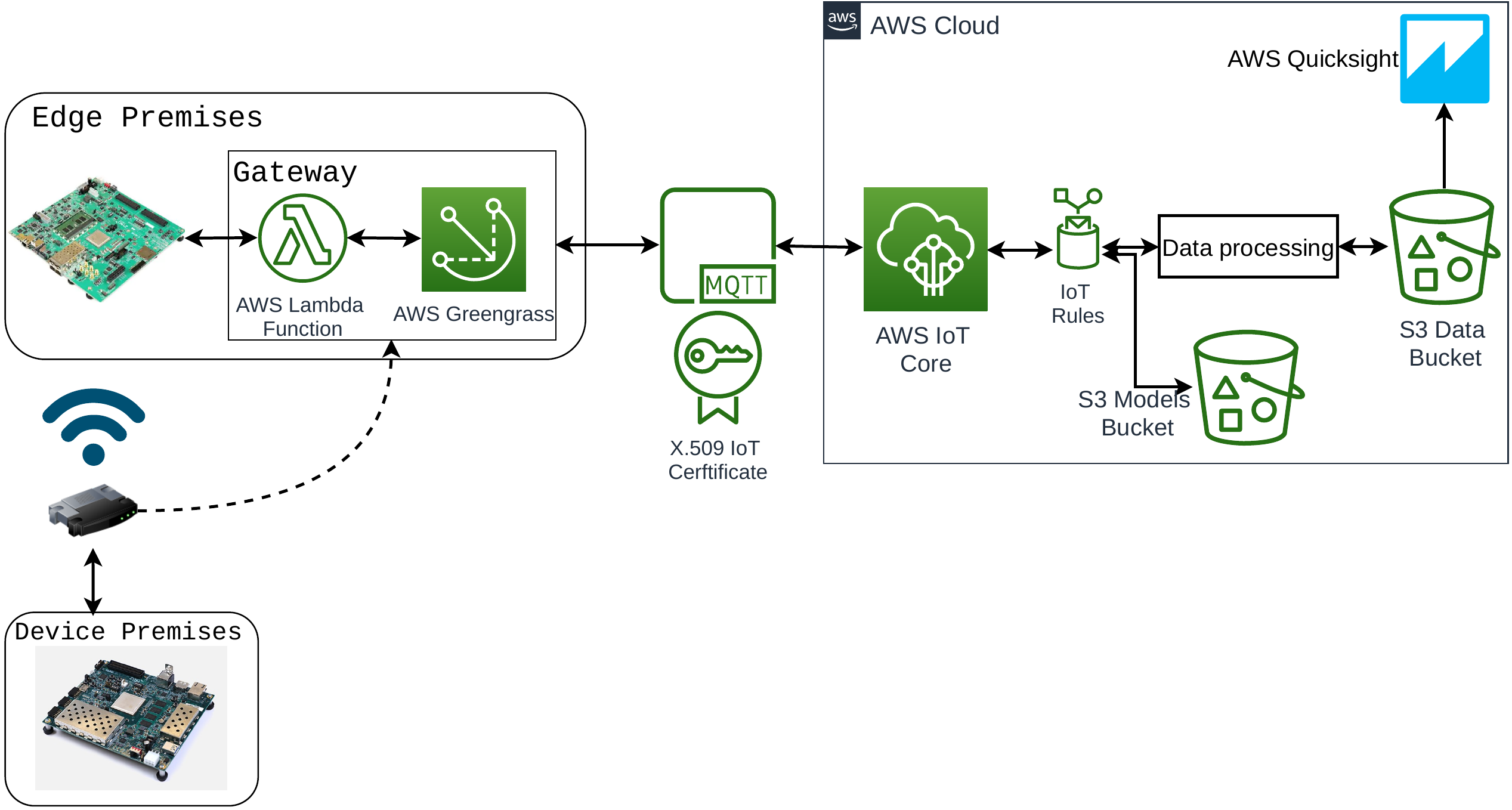}
\caption{Proposed Telemetry Infrastructure }
\label{fig:telemtery-deployed}
\vspace{-5mm}
\end{figure}
\vspace{-1mm}
\subsection{Cloud Side}
The second part of the architecture consists of the cloud side which offers services for securely receiving the telemetry data, enriching those data with additional information (e.g. timestamps), analyzing the information contained in the data, providing feedback actions to the edge platform based on some defined triggers and additional services such as further processing, storage, and analytics. 
The practical implementation used in this paper is based in the Amazon Web Services (AWS) cloud environment. 

\subsection{Telemetry Architecture}

In Figure \ref{fig:telemtery-deployed} we can see the actual components in the deployment of the telemetry architecture for both the edge and cloud sides. In the edge platform we have deployed the AWS IoT Greengrass software which provides the environment for running lambda functions to control the hardware platform and the application running on the platform. 

The workflow of the process is as follows:
At first on the edge premises the telemetry agent is running and collecting metrics from the application, AI neural network and hardware platform. The collected information is packed into a JSON object and sent to the Greengrass core (GGC) located on the edge side. The GGC is registered with AWS IoT Core on the cloud side and uses the MQTT protocol to forward the JSON objects to the cloud. The IoT Core provides a Device Gateway which manages active device connections and a powerful Message Broker which routes the messages with low latency. 
Once a message is received we use AWS IoT Rules to send messages to further data processing and aggregation before storing them to the S3 data bucket. 
Other rules are created to call specific lambda functions on the edge which perform actions like checking the achieved FPS by the object detection application and if the value is above 30 fps, lowering the clock frequency of the platform processing unit in order to save power. 
Another rule checks for model efficiency, which is the model fps divided by the ratio of peak accelerator rate and model workload, and if the number is below a certain threshold triggers a lambda function on the edge which downloads a new model from a S3 models bucket, located in the Cloud, to the edge premises to perform the inference with the new model.
In the telemetry framework in Figure \ref{fig:telemtery-deployed} the video stream is transmitted to the edge platform from the device via a 4G or 5G connection.
Beside the application, ML model, and edge platform telemetry we also collect network telemetry as explained above, which is sent to the cloud. This data can be exploited for training machine learning models able to predict the connection bandwidth from parameters collected from the router. 
There are several policies that could decide the location of the inference.
By exploiting the telemetry data collected from the router we can predict the bandwidth of the connection and decide whether it is reliable to send the video stream to the edge. There are several research work which show the possibility to predict the current connection bandwidth based on parameters like RSRP, RSRQ, and historic throughput~\cite{LinkForecast,LTENetworks,deeplearning}. In the case that the bandwidth is high enough, the stream can be transmitted to the edge premises for faster inference, otherwise the edge will decide to push the inference on the device itself, resulting in slower inference time. The decision on where to actually run the inference in this case will be made on the cloud side based on the received telemetry, and from the model results which predicts the available bandwidth. The offloading decision, from edge to device, could be made by the edge system also, which in case of high levels of utilization can decide to send the intelligent application to the device.

\section{Related Work}
There is a wide research work regarding the usage of data analytics in making smart and fast decisions especially in wireless networks~\cite{BigHealthCare,deelLearnCellNetwork,dataDriven,smallCells,MachLearResouMana}. Mainly the advances in IoT and hardware/software technology have given the opportunity to collect real-time data from user equipment or core devices which are valuable in making decisions that will impact the performance, adaptability, efficiency of the end-to-end system. %
This collection of works emphasis more the need for gathering telemetry data from different components of the end-to-end application system.
Beside the telemetry data collection there is also to consider the edge component, which in many cases is used to bring resources closer to device side and is a central actor in the real-time applications as in~\cite{fpgaRTRF, 8935405, FPGAVideoAnalytics}. 
In this paper we propose a framework that includes different telemetry data, gathered from the edge platform, application, network, and machine learning model with goal of providing feedback to the edge or device premises plus creating a data lake for training machine learning models at the cloud side. 

\section{Experimental Results}
\subsection{Latency measurements}
Devices connect to AWS IoT and other services through AWS IoT Core. Through AWS IoT Core, devices send and receive messages using device endpoints that are specific to the used AWS account.
%
\begin{comment}

\begin{figure}[h]
%
\includegraphics[width=8.5cm, height=5.5cm]{./figs/mqtt-iot-core2.pdf}
\caption{Measuring the latency of publishing data to the IoT core with MQTT on multiple messages}
\label{fig:latency-mqtt}
\vspace{-2mm}
\end{figure}

\begin{figure}[h]
%
\includegraphics[width=8.5cm, height=5.5cm]{./figs/http-iot-core-multiple2.pdf}
\caption{Measuring the latency of publishing data to the IoT core with HTTP on multiple messages }
\label{fig:latency-http-multiple}
\vspace{-7mm}
\end{figure}
\end{comment}
There are two main communication protocols for sending the data to the message broker in the IoT Core service. One is MQTT, which is a lightweight and widely adopted messaging protocol that is designed for constrained devices, and the other is HTTPS over websockets.
To evaluate the proposed architecture, we measured the achievable message latency when reaching the IoT Core and measured any possible difference between the communication protocols. 
\begin{table}[h]

\begin{center}
 \begin{tabular}{||c | c | c | c | c||} 
 \hline
 Protocol & Mean Lat.(ms) & Min(ms) & Max(ms) & Std. dev(ms) \\ [0.5ex] 
 \hline\hline
 MQTT & 516,44 & 218 & 1652 & 169,45 \\ 
 \hline
 HTTP & 565,75 & 181 & 6600 & 415,91 \\
 \hline
 \end{tabular}
\caption{Latency measurements of sending data to the cloud}
\vspace{-2em}
\label{table-latency}
\end{center}
\end{table}
All measurements were done with the Edge platform connected to a commercial 4G network, and the IoT core deployed in eu-west-2 region (London).

As shown in Table~\ref{table-latency}, for the case of MQTT the  average latency is lower regardless of the fact that with Greengrass there is an additional delay of the core software. 
Also the spikes in case of HTTP are quite high bringing a real need for local processing on the edge instead of relying only on the cloud.
%
%
%
\begin{comment}

\subsection{FPGA implementation discussion}
The Deep Learning Processing Unit (DPU) in the FPGA, offers several parameters for configuration, some of which can be modified dynamically, e.g. schedule of tasks, memory model of the DPU, frequency of the DPU core etc, and some require a FPGA reconfiguration, e.g. number of DPU cores built, number of DSP in the DPU, BRAM of the DPU etc. According to the experimental results trading off number of cores with larger block RAM for the DPU has better impact on the performance of machine vision applications. If there is the need to stay within certain power envelope of the system, increasing the schedule efficiency of the tasks in the DPU can produce beneficial results even though with reduced number of DPU cores.
The first bottleneck of the platform to have a high influence is the CPU which is handling pre/post processing tasks on the AI pipeline of the application, which can lower the platform performance. The second concern is the memory throughput available for the programmable logic which which can quickly saturate the platform availability.     
\end{comment}

\section{Conclusions and future Work}
This paper proposes an edge/cloud telemetry collection and utilization framework for applications where reliability, latency, power efficiency and high computational capacity is critical. For instance, vehicle safety as well as vehicular visual and non visual sensing systems could be potential use cases. We evaluate GPU based platform against FPGA platform for the role of edge node in an AI computer vision application and set up our framework with the FPGA platform induced by latency and power efficiency numbers provided. 
We define the cloud side components of the data lake architecture which will serve later as valuable input for training machine learning networks at the cloud side.
At the end we discuss about reaction time of cloud side of the framework and FPGA implementation issues which is good to consider when developing AI-based application on re-configurable platforms. As a future work we foresee the creation of intelligent engines on the cloud side based on the data collected through the telemetry framework. 

\section*{Acknowledgment}
This work has been partly funded by the European Commission through the projects EU-TW 5G-DIVE (Grant Agreement no. 859881).

\bibliography{references}

\begin{thebibliography}{10}

\bibitem{alibaba}
Xilinx.
\newblock Xilinx powers alibaba cloud faas with ai acceleration solution for
  e-commerce business.

\bibitem{Eyeriss}
Y.~{Chen}, T.~{Yang}, J.~{Emer}, and V.~{Sze}.
\newblock Eyeriss v2: A flexible accelerator for emerging deep neural networks
  on mobile devices.
\newblock {\em IEEE Journal on Emerging and Selected Topics in Circuits and
  Systems}, 9(2):292--308, 2019.

\bibitem{rexha2021data}
Hergys Rexha and Sebastien Lafond.
\newblock Data collection and acceleration infrastructure for fpga-based edge
  ai applications, 2021.

\bibitem{LinkForecast}
C.~{Yue}, R.~{Jin}, K.~{Suh}, Y.~{Qin}, B.~{Wang}, and W.~{Wei}.
\newblock Linkforecast: Cellular link bandwidth prediction in lte networks.
\newblock {\em IEEE Transactions on Mobile Computing}, 17(7):1582--1594, 2018.

\bibitem{LTENetworks}
N.~{Bui} and J.~{Widmer}.
\newblock Data-driven evaluation of anticipatory networking in lte networks.
\newblock {\em IEEE Transactions on Mobile Computing}, 17(10):2252--2265, 2018.

\bibitem{deeplearning}
J.~{Schmid}, M.~{Schneider}, A.~{HöB}, and B.~{Schuller}.
\newblock A deep learning approach for location independent throughput
  prediction.
\newblock In {\em 2019 IEEE International Conference on Connected Vehicles and
  Expo (ICCVE)}, pages 1--5, 2019.

\bibitem{BigHealthCare}
P.~{Jiang}, J.~{Winkley}, C.~{Zhao}, R.~{Munnoch}, G.~{Min}, and L.~T. {Yang}.
\newblock An intelligent information forwarder for healthcare big data systems
  with distributed wearable sensors.
\newblock {\em IEEE Systems Journal}, 10(3):1147--1159, 2016.

\bibitem{deelLearnCellNetwork}
J.~{Wang}, J.~{Tang}, Z.~{Xu}, Y.~{Wang}, G.~{Xue}, X.~{Zhang}, and D.~{Yang}.
\newblock Spatiotemporal modeling and prediction in cellular networks: A big
  data enabled deep learning approach.
\newblock In {\em IEEE INFOCOM 2017 - IEEE Conference on Computer
  Communications}, pages 1--9, 2017.

\bibitem{dataDriven}
L.~{Wang} and S.~{Cheng}.
\newblock Data-driven resource management for ultra-dense small cells: An
  affinity propagation clustering approach.
\newblock {\em IEEE Transactions on Network Science and Engineering},
  6(3):267--279, 2019.

\bibitem{smallCells}
L.~{Wang} and S.~{Cheng}.
\newblock Data-driven resource management for ultra-dense small cells: An
  affinity propagation clustering approach.
\newblock {\em IEEE Transactions on Network Science and Engineering},
  6(3):267--279, 2019.

\bibitem{MachLearResouMana}
F.~{Hussain}, S.~A. {Hassan}, R.~{Hussain}, and E.~{Hossain}.
\newblock Machine learning for resource management in cellular and iot
  networks: Potentials, current solutions, and open challenges.
\newblock {\em IEEE Communications Surveys Tutorials}, 22(2):1251--1275, 2020.

\bibitem{fpgaRTRF}
Z.~{Khan} and J.~J. {Lehtomäki}.
\newblock Fpga-assisted real-time rf wireless data analytics system: Design,
  implementation, and statistical analyses.
\newblock {\em IEEE Access}, 8:4383--4396, 2020.

\bibitem{8935405}
W.~{Xia}, G.~{Zheng}, Y.~{Zhu}, J.~{Zhang}, J.~{Wang}, and A.~P. {Petropulu}.
\newblock A deep learning framework for optimization of miso downlink
  beamforming.
\newblock {\em IEEE Transactions on Communications}, 68(3):1866--1880, 2020.

\bibitem{FPGAVideoAnalytics}
Shang Wang, Chen Zhang, Yuanchao Shu, and Yunxin Liu.
\newblock Live video analytics with fpga-based smart cameras.
\newblock In {\em Proceedings of the 2019 Workshop on Hot Topics in Video
  Analytics and Intelligent Edges}, HotEdgeVideo'19, page 9–14, New York, NY,
  USA, 2019. Association for Computing Machinery.

\end{thebibliography}
\bibliographystyle{unsrt}

\end{document}